\documentclass[letterpaper, 10 pt, conference]{ieeeconf}

\IEEEoverridecommandlockouts

\usepackage{color}
\usepackage{multirow}
\usepackage{booktabs}
\usepackage{url}
\usepackage{subfigure}
\usepackage{threeparttable}
\usepackage{cite}
\makeatletter
\let\NAT@parse\undefined
\makeatother
\usepackage{hyperref}[colorlinks,linkcolor=blue]
\usepackage[ruled,linesnumbered]{algorithm2e}
\renewcommand{\baselinestretch}{0.966}

\usepackage{booktabs}
\usepackage[margin=1in]{geometry}
\usepackage{array}
\newcolumntype{L}[1]{>{\raggedright\arraybackslash}p{#1}}

\usepackage{enumerate}
\usepackage{amssymb}
\usepackage{leftidx}
\usepackage{amsfonts}
\usepackage{amsmath}
\usepackage{bm}
\usepackage{booktabs}
\usepackage{setspace}
\usepackage{makecell}
\usepackage{setspace}
\usepackage{float}
\usepackage{subfigure}
\usepackage[pdftex]{graphicx}
\usepackage{cite}
\usepackage{siunitx}
\usepackage{soul}
\graphicspath{{figs/}}
\usepackage{graphicx}
\graphicspath{{figs/}}
\newcommand{\etal}{\textit{et al}.}

\newcommand{\eg}{\textit{e}.\textit{g}.}
\usepackage{tikz}
\usetikzlibrary{shapes,snakes}
\begin{document}

\title{\LARGE \bf
Online Whole-body Motion Planning for Quadrotor using Multi-resolution Search
}

\author{Yunfan~Ren*, Siqi~Liang*, Fangcheng~Zhu, Guozheng~Lu, and Fu~Zhang
 \thanks{*These two authors contributed equally to this work.}
 \thanks{Y. Ren, F. Zhu, G. Lu and F. Zhang are with the Department of Mechanical Engineering, University of Hong Kong
  \texttt{\{renyf, zhufc\}}\texttt{@connect.hku.hk},
  \texttt{\{gzlu, fuzhang\}}\texttt{@hku.hk},
 S. Liang is with School of Mechanical Engineering and Automation, Harbin Institute of Technology 
 \texttt{sqliang@stu.hit.edu.cn}.
}
}

\maketitle

\begin{tikzpicture}[overlay, remember picture]
  \path (current page.north) ++(0.0,-1.0) node[draw = black] {Accepted for 2023 IEEE International Conference on Robotics and Automation (ICRA), London, the United Kindom};
\end{tikzpicture}
\vspace{-0.3cm}

\pagestyle{empty} 
\thispagestyle{empty} 

\begin{abstract}

In this paper, we address the problem of online quadrotor whole-body motion planning (SE(3) planning) in unknown and unstructured environments. We propose a novel multi-resolution search method, which discovers narrow areas requiring full pose planning and normal areas requiring only position planning. As a consequence, a quadrotor planning problem is decomposed into several SE(3) (if necessary) and $\mathbb R^3$ sub-problems. To fly through the discovered narrow areas, a carefully designed corridor generation strategy for narrow areas is proposed, which significantly increases the planning success rate. The overall problem decomposition and hierarchical planning framework substantially accelerate the planning process, making it possible to work online with fully onboard sensing and computation in unknown environments. Extensive simulation benchmark comparisons show that the proposed method 
is one to several orders of magnitude faster than
the state-of-the-art methods in computation time while maintaining high planning success rate. The proposed method is finally integrated into a LiDAR-based autonomous quadrotor, and various real-world experiments in unknown and unstructured environments are conducted to demonstrate the outstanding performance of the proposed method.

\end{abstract}

\section{Introduction}
\label{sec:intro}

\begin{figure}[t]
 \centering 
\includegraphics[width=0.48\textwidth]{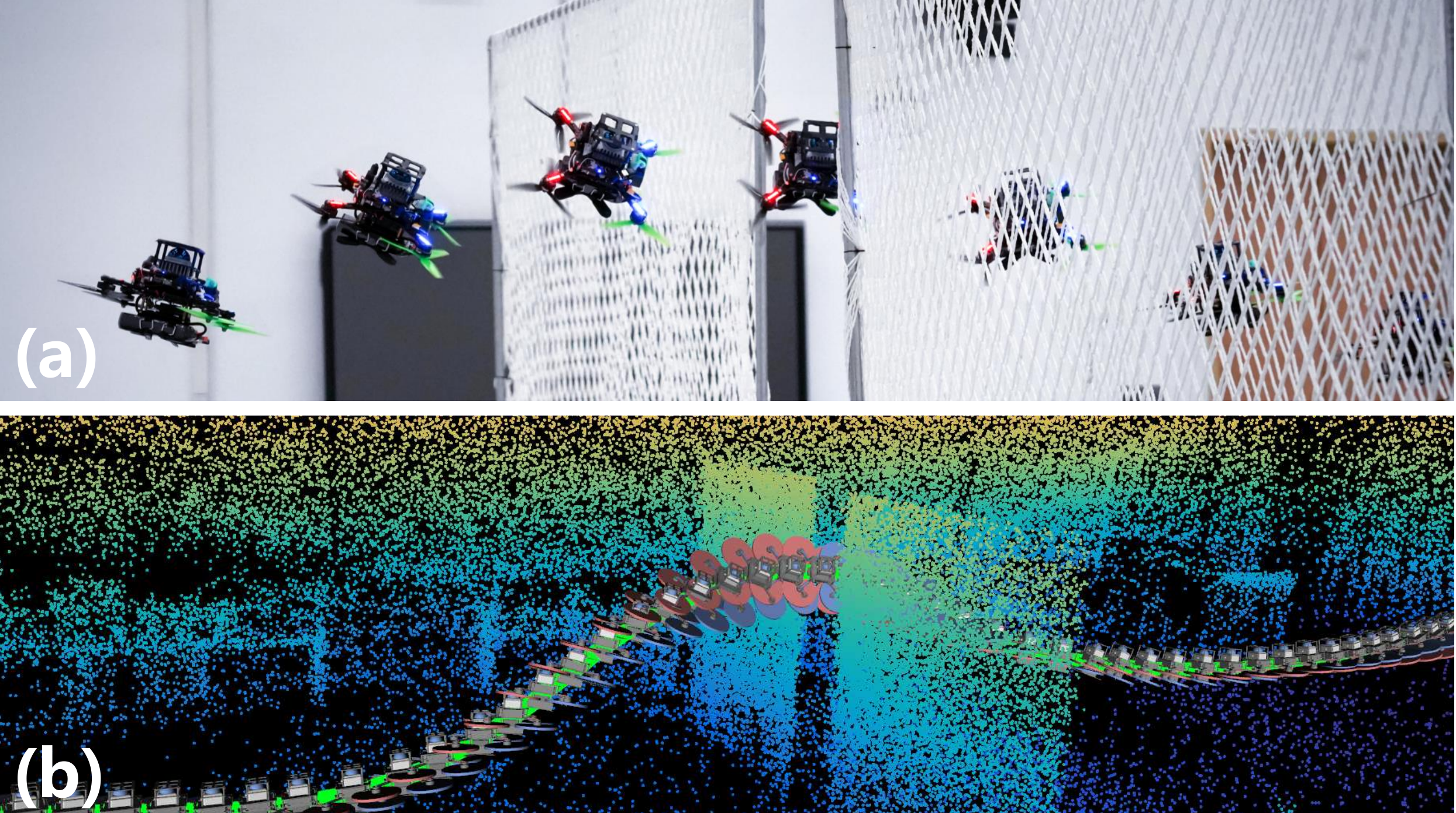}
 \caption{Performing an aggressive maneuver to fly through a narrow gap. (a) The composite image of one real-world test. (b) The point cloud view of the same flight. The trajectory is planned online without any prior information of the environment, including the narrow gap. All sensing, localization, planning, and control are running on the onboard computer in real-time.}
 \label{fig:cover}
\end{figure}

Quadrotors with a large thrust-to-weight ratio can perform extremely aggressive maneuvers, enabling applications like searching and rescuing in highly complex unstructured environments. With the whole-body motion planning (SE(3) planning), which simultaneously considers the position and attitude of a drone, quadrotors can fly through areas that are smaller than the robot's body in normal flights (\eg, narrow gap shown in Fig.~\ref{fig:cover}). 

Traditional motion planning for quadrotor like \cite{tordesillas2021faster,zhou2020ego,ren2022bubble,zhou2020robust, liu2017search} simply ignore the shape and orientation of the drone, by planning a position trajectory ($\mathbb R^3$ planning). Although $\mathbb R^3$ planning is computationally efficient, they are over-conservative since they prohibits some trajectories which are really feasible if the drone is at a certain attitude. 

Some existing works consider the drone's attitude (SE(3) planning). However, two main challenges remain to be addressed: \textbf{(1) Planning in unknown and unstructured environments}: some existing works either make a strong assumption about the environment (\eg, \cite{loianno2016estimation,hirata2014optimal,lin2019flying,yang2021whole,wang2022geometrically} assume the shape of small gaps is known or have specific visual feature) or have a low success rate in unknown, cluttered environment \cite{han2021fast}, which are not suitable for real-world applications.
 \textbf{(2) Computational efficiency}: Although some methods like \cite{han2021fast,liu2018search} can be modified to accommodate online planning, they take hundreds of milliseconds to a few seconds to compute, which limit their application in real-world online quadrotor navigation.

To address above challenges, in this paper, we propose a multi-resolution search method, which actively discovers narrow areas requiring certain drone attitude and large areas that impose no such requirement. Distinguishing narrow areas from large areas allows to decouple SE(3) planning, which is very time-consuming, from normal $\mathbb{R}^3$ planning, which is much more efficient, thus reducing the overall computation time significantly. Moreover, the awareness of narrow areas allows more dedicated planning around that area, which also improves the success rate. Requiring no prior information (\eg, position, pose, shape) of the narrow area, our method is applicable to general unknown and unstructured environments. 
To sum up, the contributions of this paper are listed below:

\begin{itemize}
 \item [1)] 
 We propose a novel parallel multi-resolution search (MR-search) method which actively discovers narrow areas requiring certain drone attitude and large areas that impose no such requirement. 
 
 \item [2)] Based on the multi-resolution search, we propose a hierarchical planning framework that separates the SE(3) planning from the $\mathbb{R}^3$ planning. 
 \item [3)]
 We present a compact LiDAR-based quadrotor that is fully autonomous and integrated onboard sensing, localization, planning, and control modules. To the best of our knowledge, this is the first time a fully autonomous quadrotor can perform online whole-body motion planning and navigation in unknown and unstructured environments.
 \item [4)]
Various simulation benchmarks comparisons show the advantages of the proposed method in terms of computation efficiency and success rate over the state-of-the-art baselines. Furthermore, extensive real-world tests show the proposed method's outstanding real-world performance.

\end{itemize}

\section{Related Works}
\label{sec:related}

Collision avoidance is a key component of robot motion planning. The mainstream planning approaches modeled the robot as a sphere\cite{ren2022bubble,zhou2019robust,kong2021avoiding,zhou2020ego,tordesillas2021faster}. In this way, the configuration space (C-space) can be easily obtained by inflating the obstacles by the largest axis length of the robot and then simplifying the drone into a mass-point and planning in the C-space (i.e., $\mathbb R^3$ planning). However, this leads to conservatism by prohibiting some trajectories that are feasible if the attitude is right. Planning both the robot's attitude and position is commonly referred to as whole-body planning, or SE(3) planning, which is necessary to navigate a robot in narrow, cluttered environments\cite{liu2018search}.

One problem of whole-body motion planning is distinguishing narrow areas requiring full pose planning. \cite{lin2019flying,wang2022geometrically} focus more on the planning and control problem while the pose and shape of the narrow gap in the environment are priorly known. Falanga \etal\cite{falanga2017aggressive} use an onboard fish-eye camera to detect the narrow gap. However, it requires the gap to have specific visual features (\eg, black and white lines) and pre-known shapes; hence cannot apply to general unstructured environments.
Our proposed method uses a novel multi-resolution search strategy to distinguish narrow areas requiring SE(3) planning from normal areas. The proposed method does not require any instrumentation of the environment nor the specific shape of the narrow area, thus applicable to general unstructured environments.

Another problem is the trajectory generation. Liu \etal~\cite{liu2018search} proposed a search-based method for quadrotor SE(3) planning. They generate a set of motion primitives and perform a graph search. To select a feasible primitive, they model the drone to an ellipsoid, then perform a collision check by verifying if any ellipsoid on the trajectory intersects with obstacles. While requiring no instrumentation of the environment or specific shape of the narrow area similar to our method, this approach cannot guarantee finding a feasible solution due to the limited discretization resolution of the motion primitive library. Also, it takes several seconds to find a trajectory, preventing this approach from an online (re-)planning framework. \cite{yang2021whole,han2021fast,wang2022geometrically} follow a two-step pipeline: they first perform convex decomposition to represent the free space by a series of convex polyhedrons. Then take the polyhedrons as the geometry constraints of the optimization problem to ensure whole-body collision avoidance. 
Those methods are one to several orders of magnitude faster than 
\cite{liu2018search} in computation time. However, they still need hundreds of milliseconds (considering the computation time of both steps) to generate a trajectory due to the large number of constraints brought by considering the robot's attitude. Moreover, without distinguishing the narrow areas, these methods' success rates are not satisfactory, as they often generate infeasible corridors in narrow areas.
In contrast, our proposed method actively discovers narrow areas and thus can perform a dedicatedly designed corridor generation strategy in narrow areas, significantly increasing the planning success rate. Furthermore, the active discovery of narrow areas makes it possible to decompose of the time-consuming SE(3) planning from normal $\mathbb R^3$ planning without planning the entire trajectory in SE(3) \cite{yang2021whole,han2021fast,wang2022geometrically}. This hierarchical planning strategy significantly reduces the computation time. The high success rate and high computational efficiency enable the proposed method to perform online quadrotor whole-body planning in unknown and unstructured environments.

\section{Preliminaries}
\label{sec:minco}

\subsection{System Modeling and Polynomial Trajectory}
\label{sec:sys}

A quadrotor system is proved to be differential flat \cite{mellinger2011minimum} with the flat output $\sigma = [x,y,z,\psi]^T$ where $\mathbf p = [x,y,z]^T$ is the position of the quadrotor in the world frame and $\psi$ the yaw angle. Since the quadrotor yaw is decoupled, we specify the yaw angle trajectory $\Phi(t)$ as the tangent direction of $\mathbf p(t)$ such that the quadrotor is always facing forward during a flight. In this way, we only need to plan the position trajectory $\mathbf p(t)$. Define the quadrotor's state as: $\mathbf s = [\mathbf p, \mathbf v, \mathbf a, \mathbf j] \in \mathbb R^{12}$, where $\mathbf v,\mathbf a,\mathbf j$ are the associated velocity, acceleration and jerk, respectively. And the kinodynamic constraints includes: $\|\mathbf v(t)\|\leq v_{\max},\|\mathbf a(t)\| \leq a_{\max},\|\mathbf j(t)\|\leq j_{\max}$.



Following \cite{wang2022geometrically}, we adopt piece-wise polynomials to represent the trajectory:
\begin{equation}
\mathbf p(t)\!=\!
\begin{cases}
\mathbf p_1(t)= \mathbf c_{1}^T\beta\left(t-0\right) & 0 \leq t<T_1 \\
\vdots & \vdots \\
\mathbf p_M(t)\!=\! \mathbf c_{M}^T\beta\left(t-T_{\!M\!-\!1}\right) \!\!\!& T_{\!M\!-\!1} \!\leq\! t\!<\!T_{\!M}
\end{cases}
\end{equation}
where $\mathbf c_1,\dots, \mathbf c_M \in \mathbb R^{2s\times 3}$ is the coefficient matrix of the $M$ pieces, and $\beta(t) = [1,t,t^2,\dots,t^{2s-1}]^T$ is the time basis vector. The state of the quadrotor can be easily obtained by taking derivative of the polynomial trajectory. In this paper, we use $s = 4$ to ensure the continuity up to jerk in adjacent pieces.

\subsection{Safety Constraints}

To avoid collisions, we use safe flight corridors (SFC) (which is a series of convex polyhedrons representing the free space) as the explicit spatial constraints. Each polyhedron is described by its $\mathcal H$-representation \cite{devadoss2011discrete}:

\begin{equation}
 \mathcal P = \{\mathbf x\in \mathbb R^3 | \mathbf A \mathbf x\preceq \mathbf b\}
\end{equation}

Each piece of a polynomial is constrained in one polyhedron. For different planning missions, we introduce two types of corridor constraints.

\begin{figure}[htbp]
 \centering 
 \includegraphics[width=0.4\textwidth]{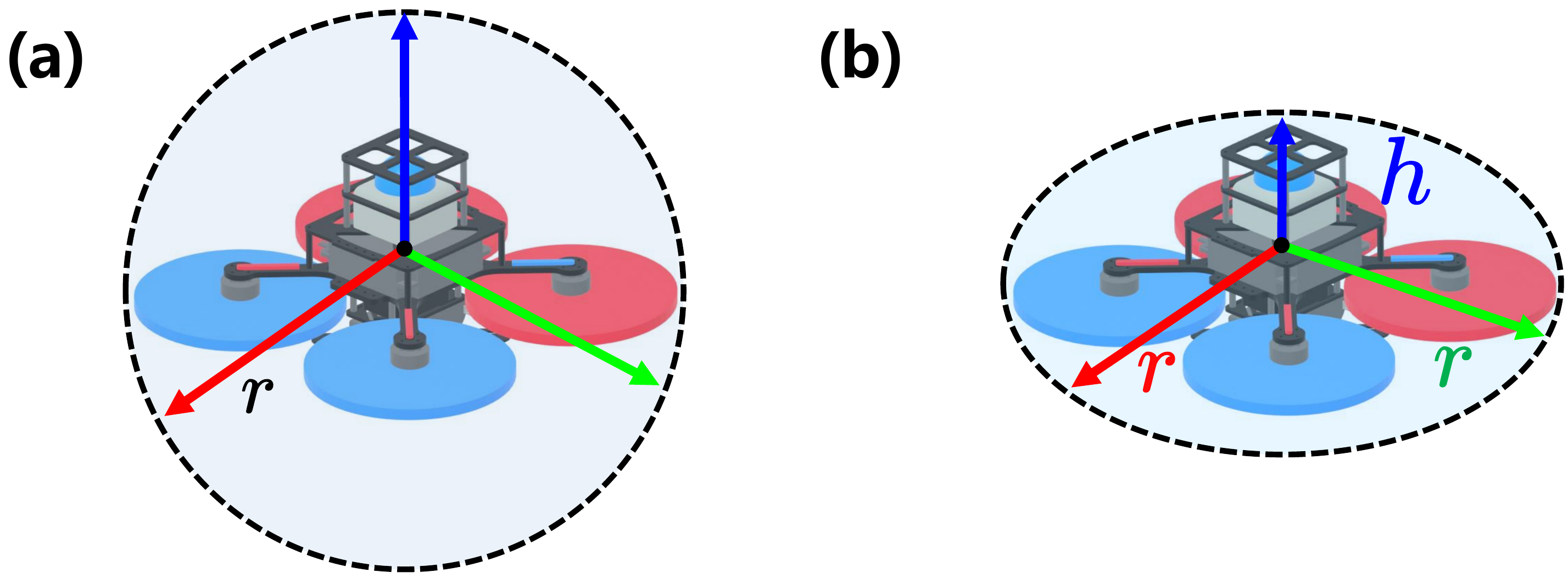}
 \caption{(a) The quadrotor is modeled as a sphere with radius equals to the largest axis length $r$ in $\mathbb R^3$ planning. (b) The quadrotor is modeled as an ellipsoid in SE(3) planning. }
 \label{fig:ellip}
\end{figure}

\subsubsection{Corridor constraints for $\mathbb R^3$ planning}
\label{sec:sfc_r3}
For $\mathbb R^3$ planning, we model it to a sphere with a radius equal to its largest axis length $r$ (see Fig.~\ref{fig:ellip}(a)). Then, we can get the configuration space by inflating all obstacle points with $r$. Finally, the corridors are generated in the configuration space. In this way, the corridor constraints for the $i$-th piece can be written as:
\begin{equation}
\label{eqa:sfc_r3}
\begin{aligned}
&{\mathbf{A}_i \mathbf p(t)- \mathbf b_i \preceq \mathbf{0}}, \forall t \in\left[T_{i-1}, T_i\right]
\end{aligned}
\end{equation}

\subsubsection{Corridor constraints for SE(3) planning}
\label{sec:sfc_se3}
 For SE(3) planning, We model the geometrical shape of the quadrotor as its outer John ellipsoid (see Fig.~\ref{fig:ellip}(b)) following \cite{wang2022geometrically}:
\begin{equation}
 \mathcal{E}(t)=\left\{\mathbf{R}(t) \mathbf{Q} \mathbf x+ \mathbf p(t) \mid\|\mathbf x\|_2 \leq 1\right\},
\end{equation}
where $(\mathbf R(t), \mathbf p(t))$ is the drone pose computed from the state $\mathbf s(t)$ based on the quadrotor differential flatness \cite{mellinger2011minimum} and
$\mathbf Q = \mathbf{diag}(r,r, h)$
is the shape matrix where $r$ is the radius of the drone and $h$ is half the height of the UAV.

As described in \cite{wang2022geometrically}, the ellipsoid along the $i$-th piece polynomial trajectory $\mathbf p_i$ inside the $i$-th polyhedron $\mathcal P_i$ can be expressed as 

\begin{equation}
\label{eqa:sfc_se3}
\begin{aligned}
&{\left[\left[\mathbf{A}_i \mathbf{R}(t) \mathbf{Q}\right]^2 \mathbf{1}\right]^{\frac{1}{2}}+\mathbf{A}_i \mathbf p_i(t)- \mathbf b_i \preceq \mathbf{0}}, \\&\forall t \in\left[T_{i-1}, T_i\right],
\end{aligned}
\end{equation}
where $\mathbf 1$ is an all-ones vector with an appropriate length, $[\cdot]^2$ and $[\cdot]^{\frac{1}{2}}$ are entry-wise square and square root, respectively. 

\subsection{Trajectory Generation}
\label{sec:gene_traj}
In this paper, the $\mathbb R^3$ trajectory generation problem is described as: for a given start state $\mathbf s_s = [\mathbf p_s, \mathbf v_s, \mathbf a_s, \mathbf j_s]$ and goal state $\mathbf s_g = [\mathbf p_g, \mathbf v_g, \mathbf a_g, \mathbf j_g]$, generates a trajectory that is collision-free and satisfies all kinodynamic constraints. The collision-free condition is described in (\ref{eqa:sfc_r3}). 
Since quadrotor's $\mathbb R^3$ planning is well studied, our $\mathbb R^3$ trajectory generation method is a combination of \cite{liu_sfc} and \cite{wang2022geometrically}. We first generate a series of SFC $\mathcal S = \{\mathcal P_1,\mathcal P_2,\dots,\mathcal P_N\}$ using RILS \cite{liu_sfc}, then perform MINCO trajectory optimization\cite{wang2022geometrically} with the SFC constraints. We called this process \textbf{GenerateR3Trajectory($\mathbf s_s$, $\mathbf s_g$)}, which will be used in the sequel.

For the SE(3) trajectory generation, we modeled the quadrotor to an ellipsoid as mentioned in Sec.~\ref{sec:sfc_se3}. Our SE(3) trajectory optimization is based on \cite{wang2022geometrically}, which generates a trajectory satisfying all kinodynamic constraints and safety constraints (\ref{eqa:sfc_se3}) within a given SFC $\mathcal S$. The SE(3) trajectory generation is encapsulated as \textbf{GenerateSE3Trajectory($\mathcal S$)}.

\section{Planner}

\begin{figure}[htbp]
 \centering 
 \includegraphics[width=0.47\textwidth]{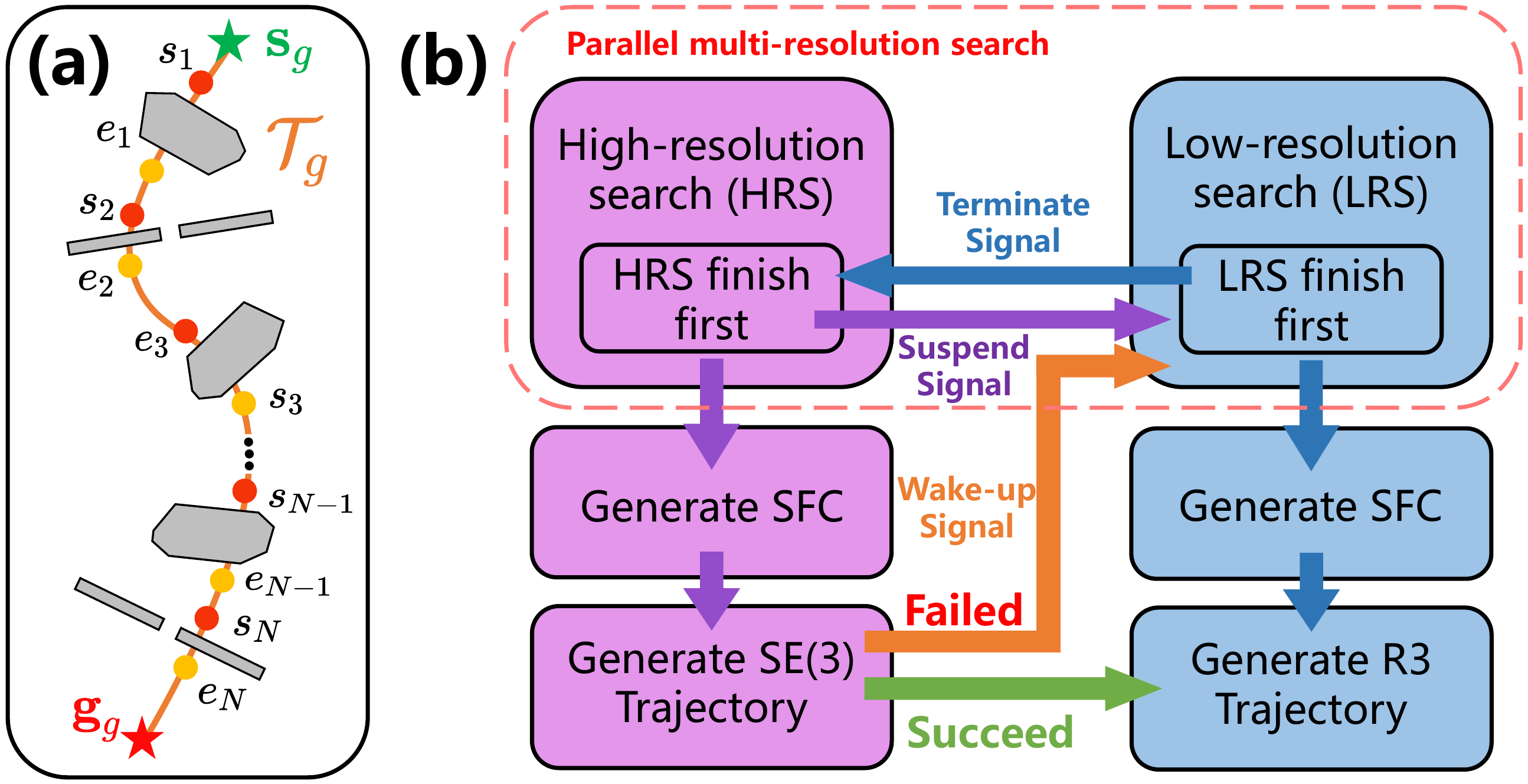}
 \caption{The overview of our proposed method.}
 \label{fig:mr_over}
\end{figure}

 The overview of the proposed method is shown in Fig.~\ref{fig:mr_over}, including: \textbf{1)} Colliding segments extraction for sub-problem generation (Sec.~\ref{sec:segments}); \textbf{2)} Parallel multi-resolution search for planning problem decomposition (Sec. \ref{sec:mr_search}); \textbf{3)} SFC generation and trajectory planning for SE(3) sub-problems. (Sec.~\ref{sec:se3_traj_gene});  \textbf{4)} $\mathbb R^3$ planning and trajectory stitching (Sec.~\ref{sec:r3_traj_gene}).

\subsection{Colliding Segments Extraction}\label{sec:segments}

In the first step, we generate a global optimal trajectory $\mathcal T_g$ to connect the global start state $\mathbf s_g$ and the global goal state $\mathbf g_g$ (see orange curve in Fig.~\ref{fig:mr_over}(a)) without consider any obstacles by solving the \textit{Linear Quadratic Minimum Time} (LQMT) problem following \cite{mueller2015computationally}. Then a collision check process is performed to extract $N$ pairs of colliding segments $\mathbf S = \{\mathbf S_i = (s_i,e_i)|i = 1,2,\dots,N\}$, where $s_i$ is the start point and $e_i$ is the end point (see red and yellow points in Fig.~\ref{fig:mr_over}(a)). 

\subsection{Parallel Multi-resolution Search}
\label{sec:mr_search}
The key of our multi-resolution search is a dual-resolution map respectively called \textit{low-resolution map} (LRM) and \textit{high-resolution map} (HRM). The LRM's resolution is equal to the length of the quadrotor's largest axis length (i.e., $r$), while the HRM's is equal to the smallest axis length (i.e., $h$), both maps are inflated by one grid to represent the configuration space. As a result, a path in the free space of the LRM is guaranteed to be collision-free regardless of the quadrotor attitude, and a path in the free space of the HRM requires a certain attitude of the quadrotor. 

As shown in Fig.~\ref{fig:mr_over}(b), for the segment $\mathbf S_i$, to determine whether it requires SE(3) planning (\eg, $\mathbf S_2$) or normal $\mathbb{R}^3$ planning (\eg, $\mathbf S_1$), we perform two parallel A* search, one on the high-resolution map (i.e., the high-resolution search (HRS)), and the other on the low-resolution map (i.e., low-resolution search (LRS)), both from $s_i$ to $e_i$.
If the LRS completes first, this segment is marked as a $\mathbb R^3$ sub-problem, and the HRS is terminated. Otherwise, if the HRS completes first, this segment is marked as a candidate SE(3) sub-problem, and the LRS is suspended and may be waken up later if the candidate SE(3) sub-problem does not has a feasible trajectory. 

\subsection{SE(3) Trajectory Generation}
\label{sec:se3_traj_gene}
\begin{figure}[htbp]
 \centering 
 \includegraphics[width=0.45\textwidth]{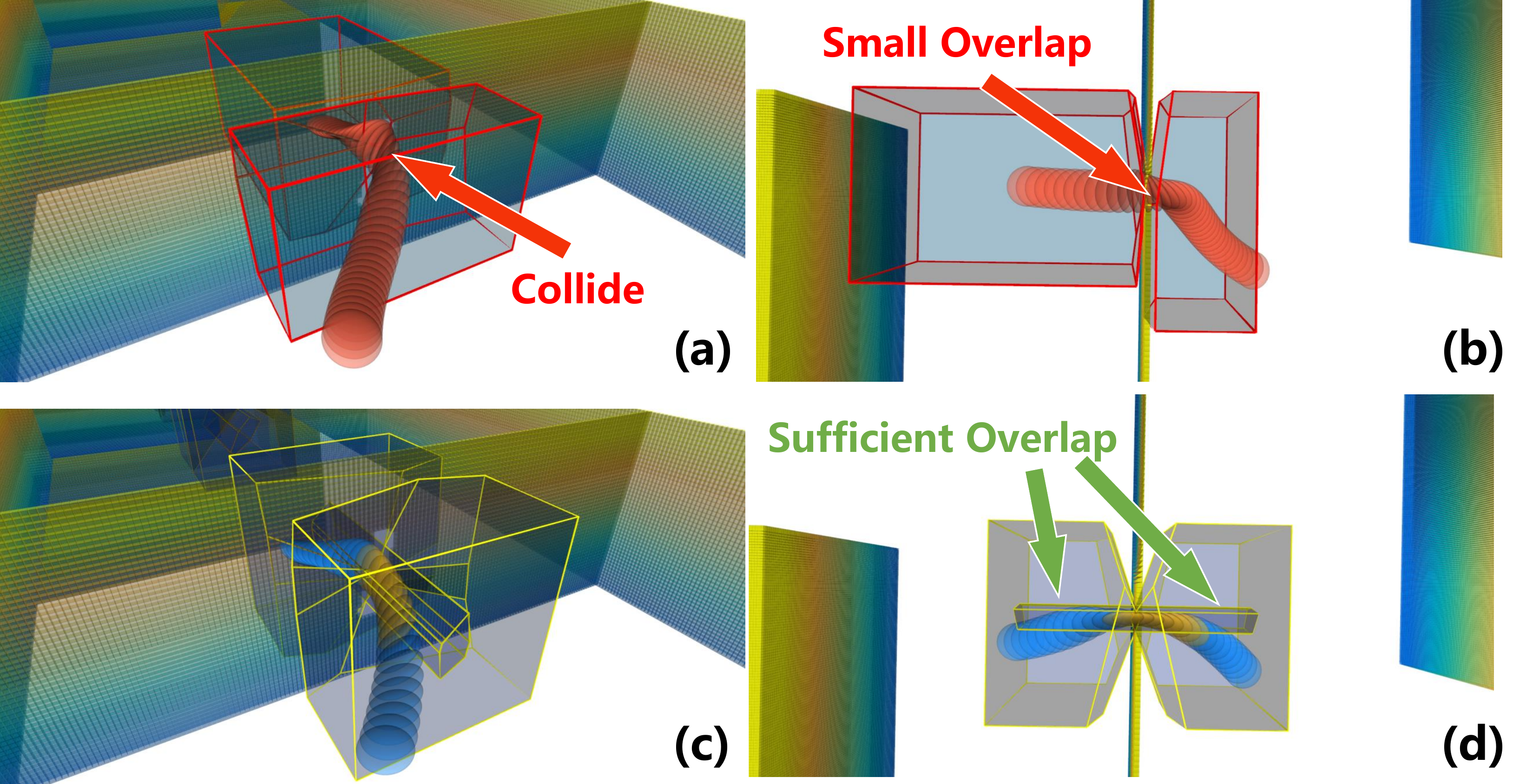}
 \caption{The generated corridor comparison. (a,b) The corridor generated by \cite{han2021fast} has two polyhedrons
 whose overlapped part cannot accommodate a quadrotor, making the optimization problem infeasible (the shown trajectory collides with the gap). (c,d) The corridor generated by the proposed method has three overlapped polyhedrons, and the shape is close to the actual free space.}
 \label{fig:sfc_cmp}
\end{figure}

If a segment $\mathbf S_i$ has been marked as a candidate SE(3) sub-problem with searched A* path $\mathcal Q$ connecting $s_i$ to $e_i$, we proceed to plan a SE(3) trajectory around the A* path. To do so, we adopt the framework in \cite{wang2022geometrically}, which is encapsulated as \textbf{GenerateSE3Trajectory($\mathcal S$)} as explained in Sec. \ref{sec:gene_traj}. A prerequisite for the algorithm is SFC $\mathcal{S}$, which consists of a series of convex polyhedrons, connecting the starting $s_i$ and end point $e_i$ of $\mathbf S_i$. However, generating convex polyhedrons around narrow areas like a small gap is not easy, existing methods like \cite{liu_sfc,han2021fast,yang2021whole} that generates SFC directly from $\mathcal Q$ often lead to small overlaps between two adjacent corridors. 
As shown in Fig.~\ref{fig:sfc_cmp}(a,b), the overlapped part is too narrow to accommodate a quadrotor, making the trajectory optimization infeasible.

\begin{figure}[htbp]
 \centering 
 \includegraphics[width=0.47\textwidth]{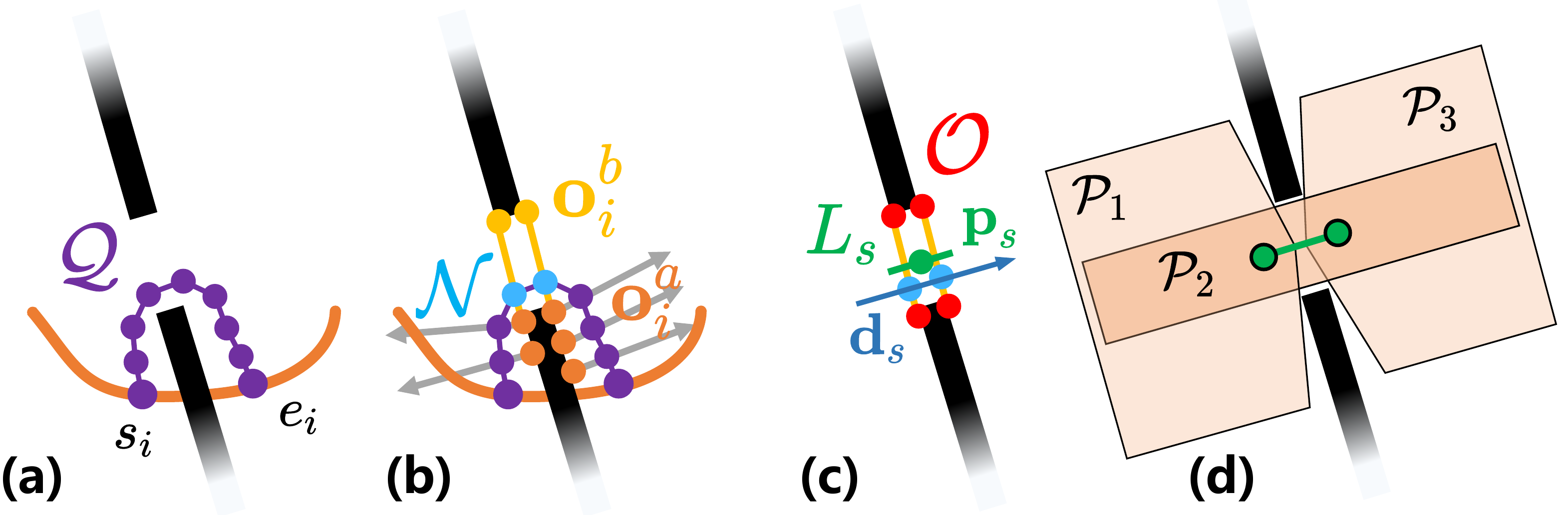}
 \caption{The proposed line seed generation processes.}
 \label{fig:se3_traj_gene}
\end{figure}

To address this problem, we present a simple seed generation method shown in Fig.~\ref{fig:se3_traj_gene}. The goal is to generate a line seed denoted by $L_s(\mathbf d_s, \mathbf p_s,l_s)$, where $\mathbf d_s$ is the direction, $\mathbf p_s$ is the midpoint and $l_s$ is the length of the line segment.
To do so, we leverage the path $\mathcal Q$ generated by the high-resolution search (see purple points in Fig.~\ref{fig:se3_traj_gene}(a)) and the high resolution map (HRM) without any inflation. For each waypoint $\mathbf q_i\in \mathcal Q$, we find its nearest occupied grid on the HRM $\mathbf o^a_i$ (see orange points in Fig.~\ref{fig:se3_traj_gene}(b)). Then, we search the first occupied grid on the HRM $\mathbf o_i^b$ (see yellow points in Fig.~\ref{fig:se3_traj_gene}(b)) along the direction of $\mathbf q_i - \mathbf o^a_i$ (see gray arrows in Fig.~\ref{fig:se3_traj_gene}(b)) and within distance $d_c = 2 r$ from $\mathbf q_i$. If such point $\mathbf o_i^b$ is found, $\mathbf o_i^b$ and $\mathbf o_i^a$ are added to $\mathcal O$, the set of points on the narrow area obstacle (see red points in Fig.~\ref{fig:se3_traj_gene}(c)), and $\mathbf q_{i}$ is added to $\mathcal N$, the set of path waypoint passing through the narrow area (see blue points in Fig.~\ref{fig:se3_traj_gene}(b)). If no such point $\mathbf o_i^b$ is found, we proceed to the next point on $\mathcal{Q}$. After this process, the line direction $\mathbf d_s$ is the average direction of $\mathcal N$ (see blue arrow in Fig.~\ref{fig:se3_traj_gene}(c)) and the midpoint $\mathbf p_s$ is center of $\mathcal O$ (see the green point in Fig.~\ref{fig:se3_traj_gene}(c)). The length of the line segment $l_s$ is set to the length of $\mathcal N$ when projected to $\mathbf d_s$. The found line seed $L_s$ is depicted by the green line in Fig.~\ref{fig:se3_traj_gene}(c). 

With the found line seed $L_s(\mathbf d_s, \mathbf p_s,l_s)$, three high-quality corridors $\mathcal S = \{\mathcal P_1,\mathcal P_2,\mathcal P_3\}$ can be generated by RILS \cite{liu_sfc} on the HRM without any inflation as shown in Fig.~\ref{fig:se3_traj_gene}(d). The $\mathcal P_1$ and $\mathcal P_3$ are respectively generated by taking the two endpoints of the $L_s$ as point seeds, and $\mathcal P_2$ passing through the small gap is generated by taking $L_s$ as a line seed. Then we perform trajectory optimization with \textbf{GenerateSE3Trajectory($\mathcal S$)} as described in Sec.~\ref{sec:gene_traj}. If the optimization is successful (which means all safety and dynamic constraints are satisfied), this segment will be marked as an SE(3) segment; If not, this segment is marked as a $\mathbb R^3$ sub-problem, and the LRS is waken up, which will continue to finding a detour path in LRM.

\subsection{$\mathbb R^3$ Trajectory Generation}
\label{sec:r3_traj_gene}
\begin{figure}[htbp]
 \centering 
 \includegraphics[width=0.45\textwidth]{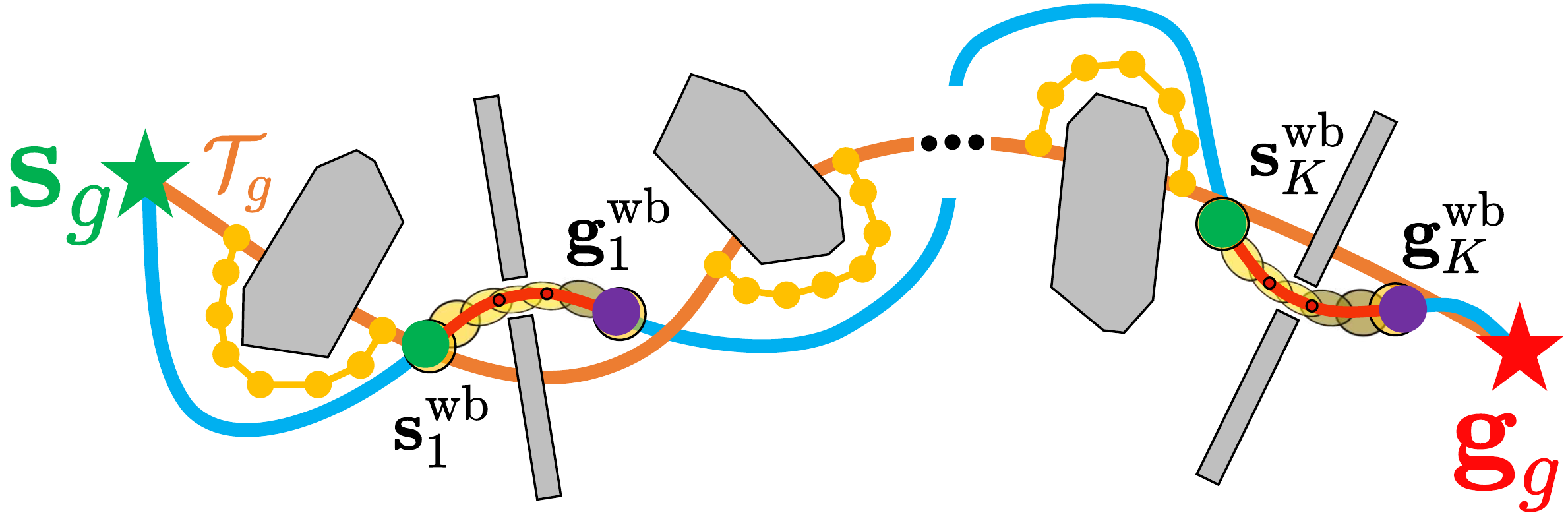}
 \caption{The blue curves represent the $\mathbb R^3$ trajectories and the red curves represent the SE(3) trajectory. The start and end states of the $\mathbb{R}^3$ trajectory are restricted to be the same as the adjacent SE(3) trajectories, ensuring the continuity up to jerk of the final trajectory}
 \label{fig:r3_traj_gene}
\end{figure}

With the steps mentioned above, we have a set of $K$ pieces of SE(3) trajectories with start states $\mathbf s_1^{\text{wb}}, \mathbf s_2^{\text{wb}},\dots, \mathbf s_K^{\text{wb}}$ and goal states $\mathbf g_1^{\text{wb}}, \mathbf g_2^{\text{wb}},\dots,\mathbf g_K^{\text{wb}}$, respectively. Between the global start state $\mathbf s_g$ and the first SE(3) trajectory, two consecutive SE(3) trajectories, or the last SE(3) trajectory and the global goal state $\mathbf g_g$, there are $\mathbb R^3$ path searched from the LRS (yellow dotted line in Fig. \ref{fig:r3_traj_gene}) and collision-free trajectory segments from $\mathcal{T}_g$ (orange curve in Fig. \ref{fig:r3_traj_gene}), we use them as the seed to produce flight corridors on the inflated low-resolution map (LRM) and plan a smooth trajectory within the corridor. The trajectory generation process is encapsulated by \textbf{GenerateR3Trajectory}(${\mathbf s}, \mathbf g$) as explained in Sec.~\ref{sec:gene_traj}, where $\mathbf s$ is the global start state or the end state of the preceding SE(3) trajectory and $\mathbf g$ is the global goal state or the start state of the succeeding SE(3) trajectory. 

\section{Experiments}

\subsection{Benchmark Comparison}
In this section, we compare the proposed method with a search-based SE(3) planning method~\cite{liu2018search} (\textit{Liu \etal}), and a corridor optimization-based method~\cite{han2021fast} (\textit{Han \etal}). Note that for \textit{Liu \etal}'s\cite{liu2018search} method, it generates a set of motion primitives and performs a graph search on it. However, it heavily suffers from the curse of dimensionality. Only a search on the x-y plane with a fixed height (i.e., 2D search) can find a feasible solution in all of the test environments below. So we manually adjust the height of the searching plane to fit the start position, goal position, and narrow gaps.

\begin{figure}[h]
 \centering
 \includegraphics[width=0.45\textwidth]{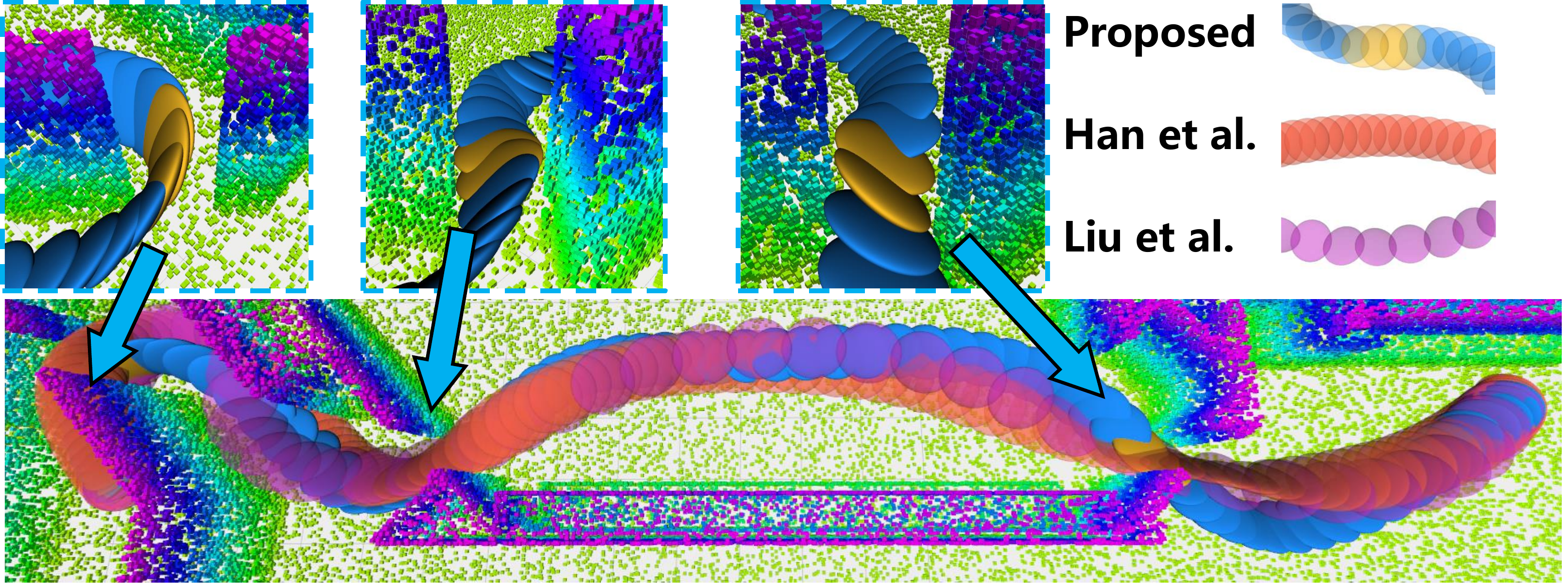}
 \caption{The generated trajectory of the three methods in \textbf{Office}.}
 \label{fig:traj_office}
\end{figure}

We make the comparison in three different simulation environments, including: \textbf{1) Office}, which is open sourced\footnote{https://github.com/sikang/mpl\_ros} from Liu \etal \cite{liu2018search}. \textbf{2) Zhangjiajie}, which is open sourced\footnote{https://github.com/ZJU-FAST-Lab/Fast-Racing} from Han \etal \cite{han2021fast}. \textbf{(3) Maze}: our custom environment. In each environment, the dynamic constraints are set to the same for all approaches.

\begin{figure}[htbp]
 \centering
 \includegraphics[width=0.45\textwidth]{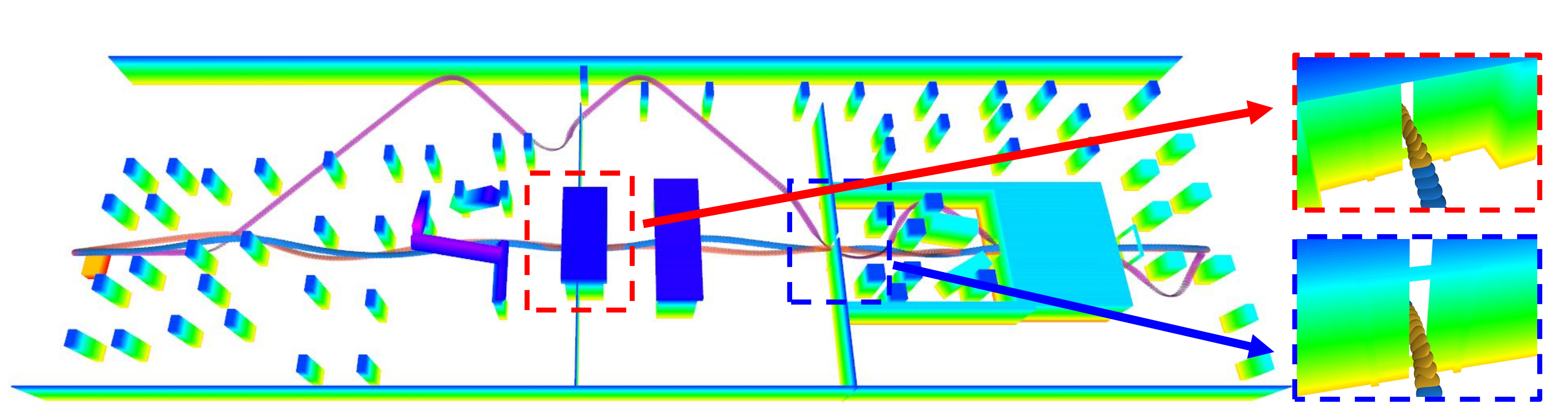}
 \caption{The generated trajectory of the three methods in \textbf{Zhangjiajie}.}
 \label{fig:traj_zhangjiajie}
\end{figure}

\begin{figure}[htbp]
 \centering 
 \includegraphics[width=0.5\textwidth]{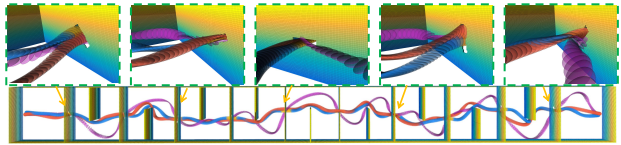}
 \caption{The generated trajectory of the three methods in \textbf{Maze}.}
 \label{fig:maze}
\end{figure}

\begin{table}[htbp]
\vspace{-0.05cm}
\footnotesize
\centering
\caption{Run Time Comparison}
\setlength{\tabcolsep}{1mm}
\begin{tabular}{@{}ccccccc@{}}\toprule
\label{tab:office}

&\multicolumn{3}{c}{Office($20.5~m$)} & \multicolumn{3}{c}{Zhangjiajie($158~m$)}\\
\cmidrule(r{4pt}){2-4} \cmidrule(l){5-7}
& $t_c$(s) & $l$ ($m$) & $t_e$(s) & $t_c$(s) & $l$ ($m$) & $t_e$(s)\\
\toprule
Liu \etal~& 1.683 & \textbf{24.80} & {5.800} & 38.210 &174.234 & 14.400
\\
Han \etal~& 0.171 &24.930 & \textbf{5.364} & 3.433 & \textbf{159.065} & \textbf{13.845}\\
Proposed & \textbf{0.0874} & 25.52 & 5.985 & \textbf{0.179} & 159.863 & {14.267}\\
\bottomrule

\end{tabular}
\end{table}

\begin{figure}[htbp]
 \centering 
 \includegraphics[width=0.47\textwidth]{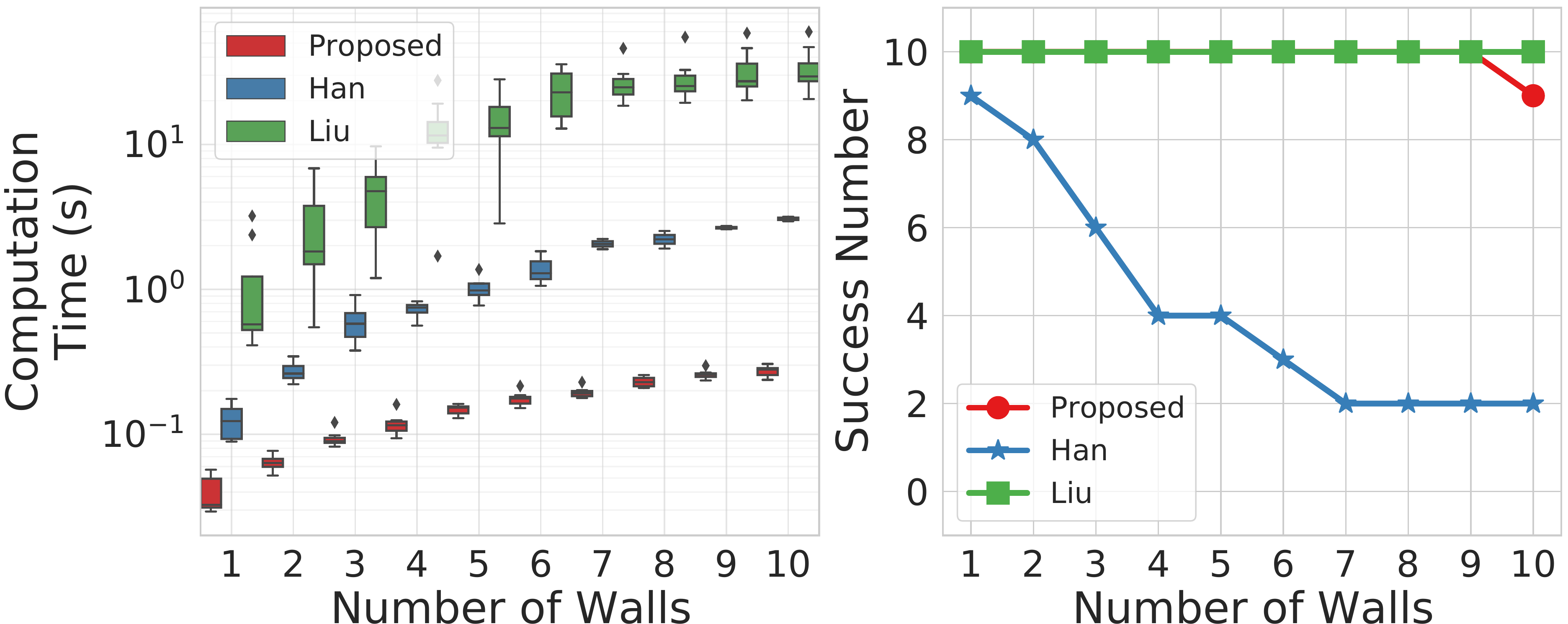}
 \caption{The computation time and success number of the three methods in \textbf{Maze}.}
 \label{fig:benchmark_maze}
\end{figure}

Since environment 1) and 2) are both static, the generated trajectory of the three methods are all the same in different trials. The generated trajectories are shown in Fig.~\ref{fig:traj_office} and Fig.~\ref{fig:traj_zhangjiajie}. 
The computation time $t_c$, trajectory tracking execution time $t_e$, and the total trajectory length $l$, are shown in Table.~\ref{tab:office} with bold font indicating better physical meaning.
The proposed method can generate trajectories with similar quality, but takes substantially less computation time. 

Our customized environment randomly generates narrow gaps on the walls with different random seeds. We set different start and goal positions making the trajectory pass through one to ten narrow gaps. For each number of walls and gaps, we tested ten times with ten different random seeds and counted the success rate. A planning is successful only if it connects the start and goal points while satisfying all dynamic constraints and ensuring collision-free. The generated trajectories are shown in Fig.~\ref{fig:maze}. And Fig.~\ref{fig:benchmark_maze} shows the computation time and success rate. As can be seen, \cite{wang2022geometrically} has a low success rate due to the reason explained in Fig. \ref{fig:sfc_cmp}. \cite{liu2018search} achieves a high success rate due to exhaustive search in a fine-grade motion library, but requires tremendously high computation time. Our proposed method is several orders of magnitude faster than the baselines while maintaining a high success rate.

\subsection{Real-world Tests}
To verify the real-world performance of the proposed method, we build a LiDAR-based quadrotor platform equipped with an Intel NUC onboard computer with CPU i7-10710U. The platform has a total weight of \SI{1.5}{\kilogram} and a thrust-to-weight ratio over $4.0$. 

High-accuracy and robust localization and mapping modules (\eg, \cite{r3live,r3live_pp,xu2022fast}) are necessary for the quadrotor to perform aggressive maneuvers.
We use the Livox Mid360 LiDAR and Pixhawk's built-in IMU running FAST-LIO2 \cite{xu2022fast}, which provides \SI{100}{\hertz} high-accuracy state estimation and \SI{50}{\hertz} point cloud. The extrinsic and time-offset between the LiDAR and IMU are pre-calibrated by \cite{zhu2022robust}. We use an on-manifold model predictive controller \cite{lu2021model} to perform high-accuracy trajectory tracking. To realize online planning and cope with newly sensed obstacles during the flight, we adopt a distance-triggered receding horizon planning scheme from our previous work \cite{ren2022bubble}. The planning horizon is set to $D=$ \SI{15}{\meter}.

\begin{figure}[htbp]
 \centering
 \includegraphics[width=0.4\textwidth]{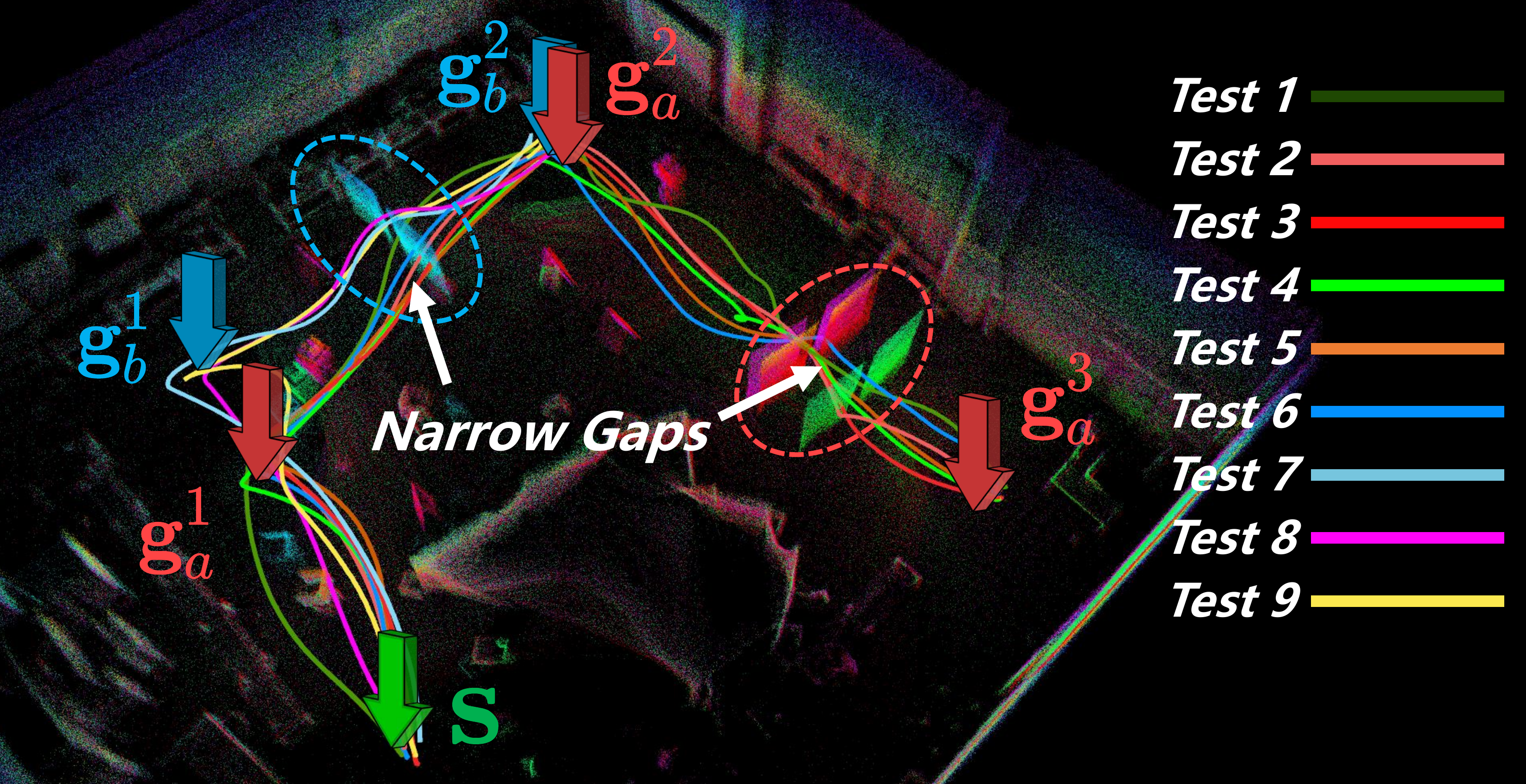}
 \caption{The composite image of the 9 tests. The point cloud is superimposed by the nine independent tests, and in each test, there is only one narrow gap. All tests start from $\mathbf s$. Test 1-6 run the path \textbf{a)}: $\mathbf s \to \mathbf g_a^1 \to \mathbf g_a^2 \to \mathbf g_a^3$, and pass through the narrow gap in red dashed ellipsoid; Test 7-9 run the path \textbf{b)}: $\mathbf s \to \mathbf g_b^1 \to \mathbf g_b^2$ and pass through the narrow gap in blue dashed ellipoid. The location of the narrow gap in each test is changed to increase the diversity. 
}
\label{fig:all_traj}
\end{figure}

\begin{figure}[htbp]
\vspace{-0.5cm}
 \centering
 \includegraphics[width=0.4\textwidth]{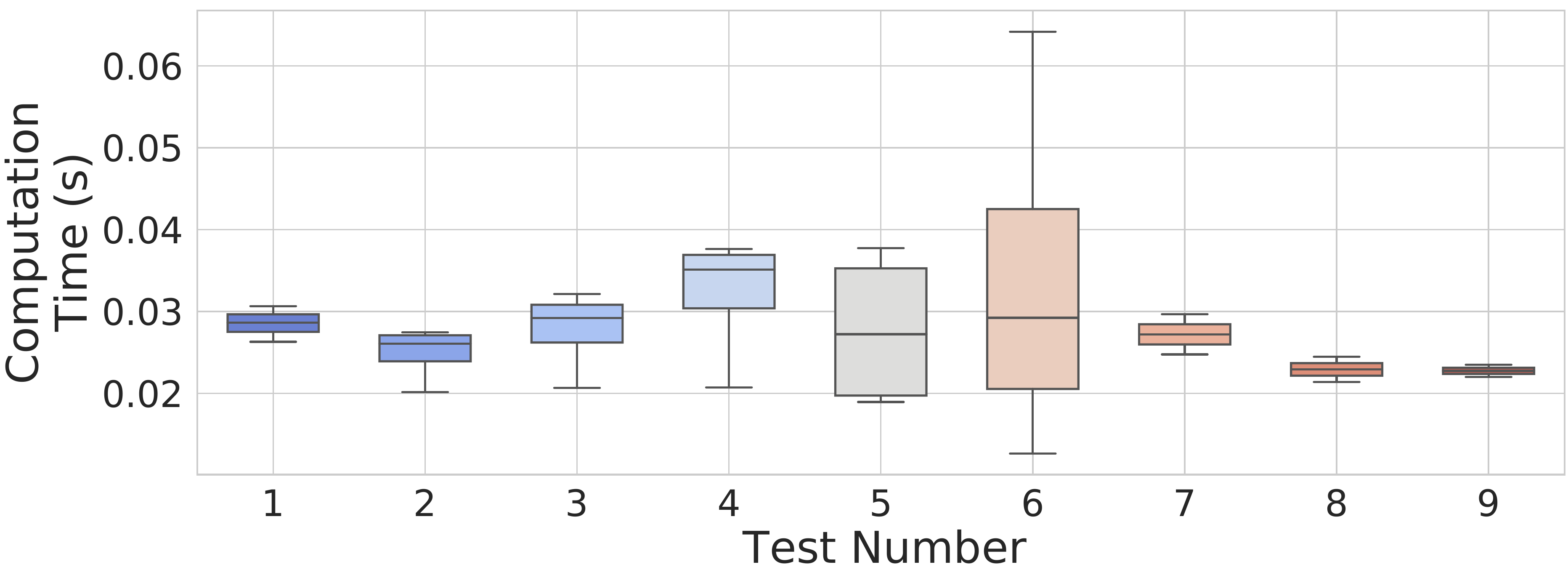}
 \caption{The computation time of each (re-)planning process in the nine real-world tests.}
 \label{fig:real_t}
\end{figure}

\begin{figure}[h]
 \centering 
 \includegraphics[width=0.43\textwidth]{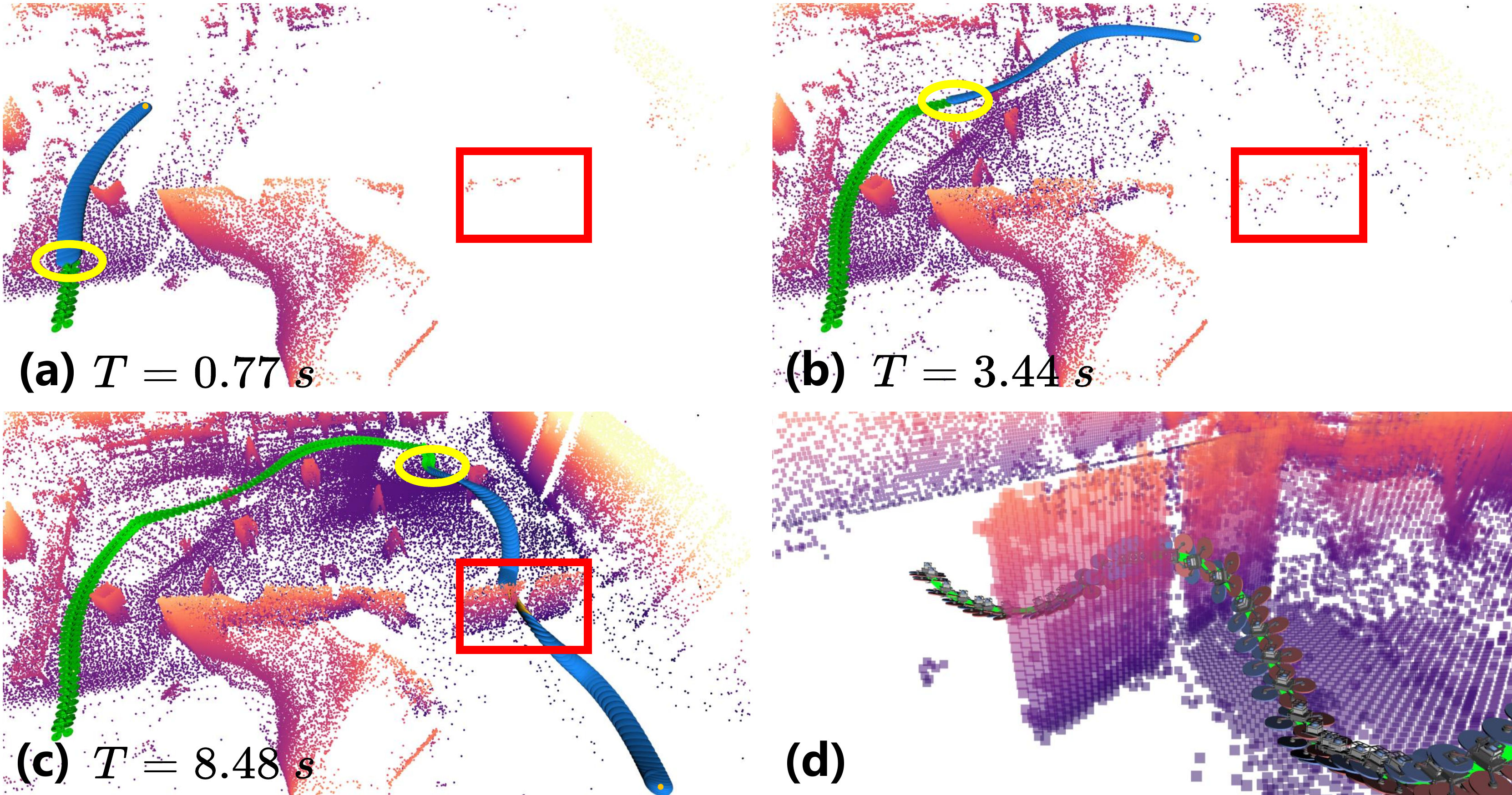}
 \caption{The point cloud during the \textit{Test 1} flight. For (a-c): the green curves indicate the executed trajectory, and the blue curves are the planned trajectory. The narrow gap (framed in the red box) is gradually sensed as the drone moves forward. The drone's position at $T = 0.77, 3.44 , 8.48$ \SI{}{s} are circled in yellow. (d) shows the trajectory of the quadrotor flying through the narrow gap.}
 \label{fig:realworld_pc}
\end{figure}

We conduct nine tests and succeed in all of them (see  Fig.~\ref{fig:all_traj}). The computation time of (re-)planning in each test is shown in Fig.~\ref{fig:real_t}, which shows the proposed method can perform over \SI{30}{\hertz} online (re-)planning with an onboard computation unit.

Due to the space limit, we only show the detail of one typical experiment called \textit{Test 1} and more details can be found in the attached video\footnote{https://youtu.be/0q5mA9vijMY}.

In \textit{Test~1}, as the drone travels forward, it automatically avoids obstacles and gradually senses the narrow gap (see Fig.~\ref{fig:realworld_pc}(a-c)). Then it plans a piece of SE(3) trajectory as shown in Fig.~\ref{fig:realworld_pc}(d). The maximum position tracking error in the middle of the gap is less than \SI{5}{cm}.

\section{Conclusion and Future Work}
In this work, we proposed an online planning framework for quadrotor whole-body motion planning in unknown and unstructured environments. We first perform a parallel multi-resolution search to decompose the planning problem into several SE(3) and $\mathbb R^3$ sub-problems. Then SE(3) problem is solved with a dedicatedly designed seed generation approach which significantly increases the success rate. The overall hierarchical planning process dramatically reduces the computation time, making it possible to perform online aggressive flights in cluttered environments with a fully autonomous drone.

One limitation of the proposed method is that the sensor can only perceive one side of the narrow area, if the narrow area is long (not a thin wall), the narrow area is not actually passable since the quadrotor cannot maintain a nonzero roll angle without inducing lateral motion. One possible direction to address this problem is to adopt perception-aware planning, which actively detects the narrow areas and reduces the influence of partial perception. In the future, we will explore these designs and extend this method to more challenging environments and missions.

\section*{Acknowledgment}
This work was supported by the Hong Kong General Research Fund (GRF) (17206920). The authors gratefully acknowledge DJI and Livox Technology for equipment support during the project. The authors would like to thank Wei Xu and Yixi Cai for the helpful discussions, and Prof. Ximin Lyu for the support of the experiment site.
\newpage
{\small
\bibliographystyle{unsrt}
\bibliography{reference}
}

\end{document}